# Multi-Module G2P Converter for Persian Focusing on Relations between Words

*Mahdi Rezaei, Negar Nayeri, Saeed Farzi, Hossein Sameti*

K. N. Toosi University of Technology, Tehran, Iran
Sharif University of Technology, Tehran, Iran

```
mahdirezaey1999@email.kntu.ac.ir, negar.nayeri@alum.sharif.edu,
       saeedfarzi@kntu.ac.ir, sameti@sharif.edu
```

## Abstract

In this paper, we investigate the application of end-to-end and multi-module frameworks for G2P conversion for the Persian language. The results demonstrate that our proposed multi-module G2P system outperforms our end-to-end systems in terms of accuracy and speed. The system consists of a pronunciation dictionary as our look-up table, along with separate models to handle homographs, OOVs and *ezafe* in Persian created using GRU and Transformer architectures. The system is sequence-level rather than word-level, which allows it to effectively capture the unwritten relations between words (cross-word information) necessary for homograph disambiguation and *ezafe* recognition without the need for any pre-processing. After evaluation, our system achieved a 94.48% word-level accuracy, outperforming the previous G2P systems for Persian.

## 1. Introduction

Before aiming to produce a Grapheme-to-phoneme (G2P) system in the current research, we attempted to create a Tacotron to receive grapheme sequences and return speech. However, the model achieved poor results due to Persian being a low-resource language; hence we decided to build a G2P system. G2P systems aim to convert a grapheme (letter) sequence into its pronunciation sequence, and are an essential component of text-to-speech (TTS) and speech recognition systems for any language lacking consistent pronunciation rules.

A good G2P system must address the issues of out-of-vocabulary (OOV) words and cross-word relations. OOV words are those which are not present in the lexicon, meaning they were not seen during model training. In the case of G2P, the lexicon is a dictionary consisting of graphemes and their respective phonemes. As for cross-word relations, a Persian G2P task is mainly concerned with homographs and *ezafe* constructions.

Homographs are words that are represented by the same grapheme, but are pronounced differently. Homograph detection in Persian is much more complicated than in English since there are more homographs in Persian vocabulary and more possible pronunciations for each one. The difference in the part-of-speech (POS) tags of homographs can sometimes help disambiguate the pronunciation of these words; however, it is not an inclusive solution since some homographs also have the same POS tag. For instance, the Persian grapheme "می‌برد" can maintain three meanings: "is taking", "is cutting" and "was taking", all verbs, and respectively pronounced [mi:bæræd], [mi:boræd] and [mi:bord]. Here, the difference in pronunciation is because, in many cases in Persian, vowels such as [æ], [e] and [ɔ] are represented by diacritics, which are often absent in Persian text. This absence of representation is among the reasons why there is usually no one-to-one relationship between graphemes and phonemes in Persian.

One of the significant tasks while processing Persian text is *ezafe* recognition. *Ezafe* links two words and creates compounds such as nominal and possessive. However, *ezafe* does not occur *whenever* a noun is followed by an adjective or another noun, thus making context information a necessary factor in *ezafe* recognition. Table 1 shows two example sentences along with their pronunciations and English translations. The highlighted words in Example 1 have an unrepresented *ezafe* in between that links them, whereas this is not the case for Example 2.

Table 1. Impact of context on *ezafe* formation.

| Example | | Phonetics | English Translation |
|---|---|---|---|
| 1 | خودکار قرمز را برداشتم. | [xodkɑ:re ʁermez rɑ: bærdɑ:ʃtæm] | I picked up the red pen. |
| 2 | رنگ آن خودکار قرمز است. | [rænge ʌn xodkɑ:r ʁermez æst] | The color of that pen is red. |

*Ezafe* is often absent from Persian text, further complicating tasks such as G2P; however, if represented explicitly, it appears with the following graphemes: a) with the diacritic "◌ِ" called *kasre* and pronounced [e], used when the first word in the compound ends in a consonant, and b) with "ی" or "ـهٔ", both pronounced [je], used when the first word in the compound ends in a vowel. Table 2 offers examples of *ezafe* representations followed by their pronunciation and English translation.

Table 2. Examples of *ezafe* representations in Persian text.

| *Ezafe* Compound | Phonetics | English Translation |
|---|---|---|
| دفترِ آبی | [dæftære ɑ:bi] | blue notebook |
| خانه‌ی ما | [xɑ:neje mɑ:] | our house |
| خانهٔ ما | [xɑ:neje mɑ:] | our house |

Considering the above-mentioned problems, we have questioned using an end-to-end approach for our G2P system. After several experiments, we concluded that assigning separate modules to each problem would increase the accuracy as well as the performance of the system since the modules can run simultaneously. Moreover, by using a pronunciation dictionary, we could reach 100% accuracy in predicting the pronunciation of many grapheme sequences. It is worth mentioning that we used no pre-processing in terms of splitting words into their composing morphemes.

Cross-word relations exist in different languages in various forms and if not represented in text, can lead to complexities in G2P and incorrect pronunciations. For example, in Arabic, different grammatical relations, POS of words and definiteness of nouns can assign certain diacritics to the last character of words in sentences. There are more diacritics in Arabic than in Persian; however, as is the case in Persian, they are often absent from the text. Additionally, in languages such as Russian and Thai, a change in the position of stress in words can change the meaning or produce a meaningless sentence [1], [2].

In Persian text, many letters have an initial as well as a final form. For example, "بـ" and "هـ" are the initial forms of the letters *be* and *he*, while their final forms are represented as "ب" and "ه". The zero-width non-joiner (ZWNJ) character (with the Unicode symbol U+200C) is placed between a letter in its final form and another in its initial form. The token "کتاب‌هایشان", meaning "their books", contains a ZWNJ between letters

ب and ه. As seen in this example, the ZWNJ character can produce syntactically-loaded tokens. We can add such tokens to the Persian training lexicon to enrich the pronunciation dictionary and increase G2P performance.

In this paper, we review related works and state-of-the-art approaches addressing the problem of G2P conversion (section 2) and introduce our proposed method to solve this problem in Persian (section 3). We then compare our experiments with previous works (section 4) and evaluate and compare the performance of our models (section 5). Finally, we propose recommendations for future work (section 6) and draw conclusions (section 7).

## 2. Related Works

The main solutions proposed to solve the G2P problem include rule-based models, joint-sequence models [3], [4], deep neural networks, and as of recent, Transformer-based models.

Joint-sequence models create graphonemes through generating alignments between graphemes and phonemes and build an n-gram model, which returns the probability of a graphoneme given *n* previous graphonemes. [3] constructs a joint-based G2P converter for Portuguese with embedded rules for stressed vowel assignment, using a pronunciation dictionary containing the most frequent words alongside a list of homographs.

[5] trained three sets of Transformers with different structures and numbers of encoder and decoder layers. The fewer parameters in the proposed method compared to previous DNN models with convolutional and recurrent layers led to faster training, while achieving a similar accuracy. The models were evaluated on PER and WER. [6] experimented with sequence and token level knowledge distillation using single and ensemble teacher models to label words with their phoneme sequences. Their proposed Transformer-based model outperformed the CNN and RNN-based ones for the converter model. This model used token-level ensemble distillation to increase the task's accuracy and improve WER and PER. Knowledge transfer was also performed to achieve lightweight models for better online deployment.

The models mentioned above receive words in isolation without considering the context information, which would significantly decrease the performance for languages with complex cross-word relations such as Persian. Moreover, G2P models have often been evaluated on CMUDict dataset for US English [5], [6], [7], [8] and NetTalk [5], [8], making them directly comparable. This comparison is not applicable to languages other than English. In the following, we review [9], [2] and [7] as examples of research that pay attention to cross-word information.

[9] built an end-to-end framework to receive raw Chinese text and perform polyphone disambiguation with the help of a pre-trained BERT. The contextual information helped the model extract semantic features and effectively disambiguate polyphones. [2] created a unified-DNN-based G2P converter for languages with regular and/or irregular pronunciation, which was verified for English, Russian and Czech. The proposed models consisted of an embedding layer, a bi-LSTM layer, another LSTM/convolutional layer, a bi-LSTM layer for the encoder, and an LSTM decoder followed by a linear projection with *softmax* activation. The model with the LSTM variation in the encoder outperformed the one with the convolutional layer. This architecture considers cross-word and within-word relations and is basically the same architecture we used for one of our proposed systems in the present paper (section 4). [7] developed three neural networks with recurrent, convolutional and Transformer architectures, among which the Transformer obtained the best results for both word and phoneme error rates in English, Croatian and Turkish. The model handles the homograph problem better when trained on the sentence level rather than word level. As opposed to previous G2P models, the proposed Transformer model is more robust against longer graphemes. The paper mentions

that rule-based models complement neural networks to better overcome the OOV problem and return lower error rates.

[10], [11] and [12] acknowledge Persian as a challenging language for the G2P conversion task due to the absence of vowel diacritics from the text. [10] develops a BRNN-LSTM accepting 10-sentence sequences derived from FarsDat and Bijankhan corpora. The model achieved a 98% accuracy on the phoneme level. [10] briefly mentions the multi-module approach—which is thoroughly raised and examined in the current article—to tackle homograph and *ezafe* problems in Persian but refuses to adopt the approach, arguing that the cumulative errors from the multiple modules would negatively affect the overall accuracy. [11] builds a multi-layer Perceptron (MLP) to solve this problem and defines a variable called *Vowel State*, which determines whether or not a letter contains a vowel and the type of vowel. The model gained a 97.34% accuracy on the phoneme level when tested on 2024 common Persian words. [12] proposed a G2P system containing a rule-based layer and two MLP layers to determine pronunciations and gemination, which obtained an 87% word-to-phoneme and a 61% letter-to-phoneme accuracy on a 3000-word test set. Unlike [10], [11] and [12] have not addressed the issues of *ezafe* and homographs, i.e., cross-word relations in Persian.

## 3. Proposed Work

In our proposed architecture, we take a multi-module approach to the problem and allocate OOV words, *ezafe* recognition and homographs to separate modules. This multi-module system also uses a dictionary containing words and their phoneme transcription. After normalizing and tokenizing the data, we loop through every word in the text sequence. If the word is in the vocabulary, i.e., within the dictionary, its pronunciation is extracted from the dictionary. If the word is within our homograph list, it is passed to the homograph model along with the two nearest preceding and proceeding words; otherwise, it will be passed to the OOV model. It is worth mentioning that all words go through *ezafe* recognition to determine whether they should be followed by an *ezafe*. The *ezafe* model receives the words along with their preceding and proceeding words the same way as the homograph model. Lastly, the results are concatenated and the G2P system generates the phoneme sequence. The models are described in the following.

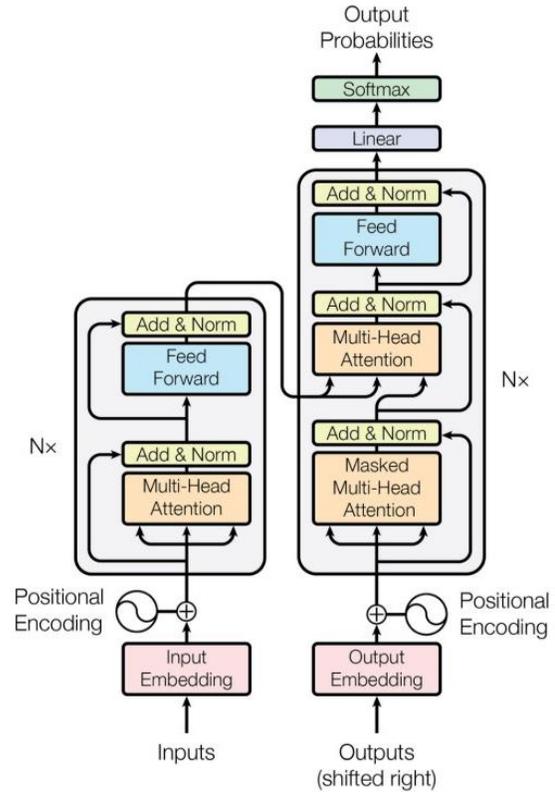

Figure 1. Transformer architecture in OOV module [13].

### 3.1. OOV Module

OOVs or out-of-vocabulary words are those which did not exist in the model's training set. Predicting the pronunciation of OOVs is a simple seq2seq task assigned to a transformer model. Figure 1 [13] shows the arch of the Transformer. The graphemes (characters) of the input words are fed into the model. Since this model is on the word level and is not concerned with cross-word relations, the phoneme sequence output does not contain any *ezafe* pronunciation at its end, which is also the case for the homograph model.

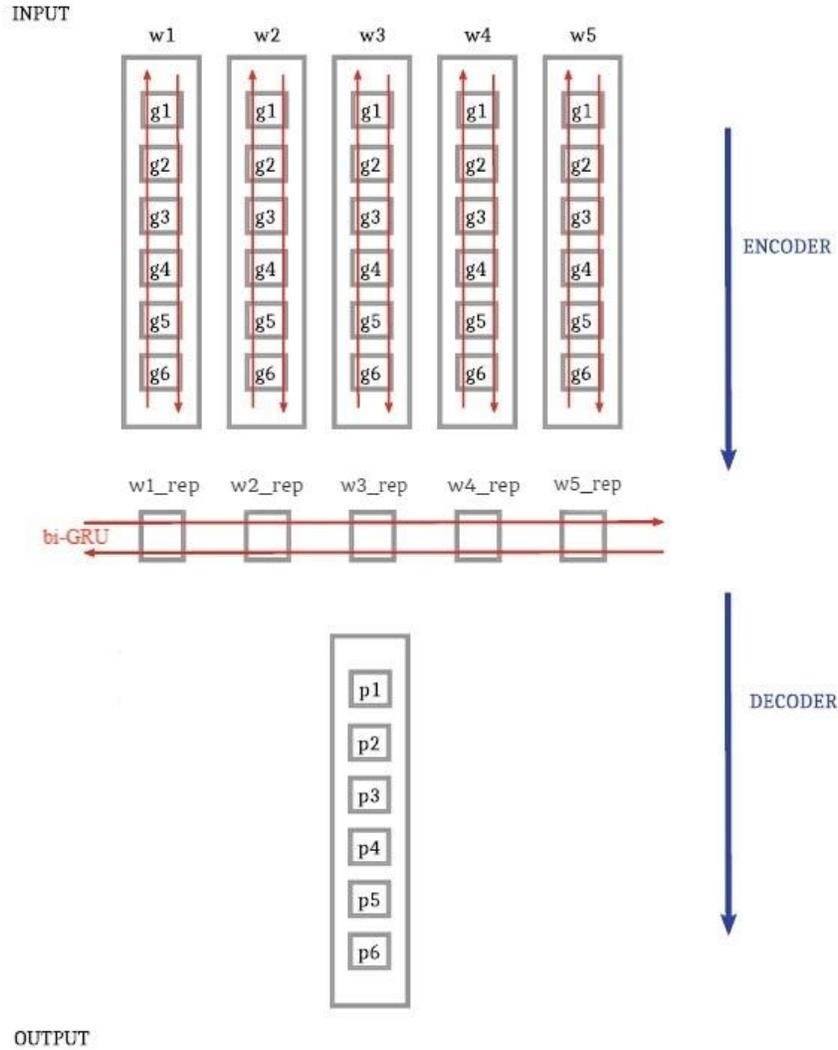

Figure 2. The structure of our bidirectional GRU neural network in the homograph model [14].

### 3.2. Homograph Module

After normalization and tokenization, similar to the architecture of the G2P converter used in [2], the text data input is fed into the homograph model in 5-gram sequences. The embedding layer performs vectorization for each character in each of the five words (tokens), and padding is applied for words shorter than the maximum length of graphemes (characters) in each batch. Five bidirectional GRU models extract within-word relations separately for each word (gram). In the next step, we create five word representation vectors using the hidden activations of those five GRU layers. We make the word representations in the same way as encoder-decoder connectors in the machine translation task. The word representations are received by another bidirectional GRU layer which is responsible for extracting cross-word information in terms of boundary and *ezafe*. Finally, the within-word and cross-word information extracted by the models regarding the middle word are concatenated and fed into a GRU decoder. Here, the cross-word information is in fact the meaning and the corresponding pronunciation of the homograph. Figure 2 derived from [14] shows the structure of the homograph model.

A homograph dictionary was extracted from the Ariana lexicon, which is introduced in 4.1. In case the model generates a phoneme sequence that is not among the pronunciations in the homograph dictionary, the sequence is corrected using a minimum edit distance algorithm, and the pronunciation with the highest score is selected. In the case where there are equal scores, one pronunciation is selected at random.

Here, it is worth mentioning the case of multi-pronunciation words, which are words with more than one correct pronunciation all having the same meaning, unlike homographs. These words are passed to our pronunciation dictionary, which only contains their most common pronunciation. For instance, the word "آسمان", meaning sky, has the two possible pronunciations, "aseman" and "asman", whereas the pronunciation dictionary only contains the former.

### 3.3. *Ezafe* Module

In one experiment, the *ezafe* module uses a deep neural network to predict whether the middle word in a 5-gram sequence has *ezafe* at its end. Very similar to the architecture in the homograph model, we create five word representation vectors, and then, thanks to a bidirectional GRU, cross-word information is extracted. This information output is passed to a linear layer rather than a decoder, which will return a 0-1 vector predicting the presence or absence of *ezafe* in the middle word in the 5-gram. This model will be called *ezafe* model I.

In another experiment, we attempted to create the *ezafe* model on the word level (*ezafe* model II). The embedding layer performed vectorization for each token in the 5-gram, leading to a large embedding table. A bidirectional GRU learned the cross-word information, which was passed to a linear layer and 0-1 vectors were returned, indicating the presence of *ezafe* in the middle word.

While labeling the training data for this model, we paid attention to pairs of grapheme and phoneme sequences due to the similar pronunciation of definite words and *ezafe* forms in Persian. For instance, "مداده" and "مدادِ" are both pronounced "*medade*", while the former means "the pencil" and the latter is pencil followed by *ezafe*.

## 4. Experiments and Results

In this section, after introducing the datasets used to train and evaluate the models, we introduce other state-of-the-art architectures for later comparison in section 5. These architectures use a single end-to-end model which simultaneously addresses issues of OOVs, *ezafe* and homographs in the same module.

### 4.1. Datasets

We used a portion of the Bijankhan corpus [15], which is the most well-known corpus for Persian G2P and consists of formal sentences to train and evaluate the G2P system. Our data consists of 42,540 sentences with an overall of one million words and their phonemic transcriptions. Our phonetic experts corrected the existing errors in the transcriptions. Furthermore, phonemes are represented with only one symbol. For example, the word "شکسته" (*shekaste*), meaning broken, is transcribed as "$ek/ste". Here, the advantage of using *$* over *sh* is that the model makes one prediction for this single phoneme rather than two, which ultimately affects the accuracy and efficiency of the model.

The data was randomly shuffled and split into train, validation and test sets with a percentage of respectively 80%, 5% and 15%, and changed into the input format required by each model. The same data was used for all modules and systems, except for OOV and homograph modules which used a subset of the original test set. This is why we used a larger portion of data for the test set compared to the validation set (15% vs. 5% of the original data).

While training the OOV model, we used all occurrences of all tokens so that the model would gain an understanding of the frequency and importance of words. We did not exclude homographs from the training set for the OOV model, and the homograph model was also trained on non-homographs as middle grams. This way, the models are trained on more data, often of the same roots, and thus, generate more phoneme sequences during training time.

The Ariana lexicon was used to tag the words in the Bijankhan corpus for the purpose of training the *ezafe* model (3.3). This lexicon consists of word entries along with all possible phoneme transcriptions and POS tags in both simple and *ezafe* forms. The Ariana lexicon was also used to create our look-up table which is a pronunciation dictionary and a list of homograph words. Figure 3 shows an example of word entries in the Ariana lexicon. The symbol GEN refers to the presence of *ezafe* in the phonemic transcription. Moreover, in Persian, some syntactic forms are attached to words with zero-width non-joiner (ZWNJ) character and form a single token, as in "کتاب‌هایشان" which means "their books". This lexicon considers a separate entry for these syntactically loaded tokens.

| نفتالین | (N1,n/ftalin)(N1GEN,n/ftaline) |
| نفتالین‌زده | (A0,n/ftalinz/de)(A0GEN,n/ftalinz/deye) |
| نفتای | (N1GEN,n/ftaye) |
| نفت‌خوار | (A0,n/ftxar)(A0GEN,n/ftxare) |
| نفت‌خواران | (A0,n/ftxaran)(A0GEN,n/ftxarane) |
| نفتش | (N1,n/ft/$) |
| نفتشان | (N1,n/fte$an) |
| نفتکش‌های | (N1GEN,n/ftke$haye) |
| نفتی‌اش | (A0,n/fti@/$) |

Figure 3. An example of entries in the Ariana lexicon.

### 4.2. Other State-of-the-Art Techniques

Based on the G2P converter developed in [2] and [14] and similar to the homograph model in 3.2, we used GRU models to learn within-word and cross-word information. The information about the middle word is then concatenated and fed into a single GRU decoder. This DNN-based model produced relatively poor results, so we did not calculate its performance and report it here. The other G2P system differed from this DNN-based system in using the attention mechanism in the decoder.

Similar to the end-to-end Transformer-based G2P converter with sentence-level inputs in [7], our Transformer-based system used an end-to-end transformer-based model which received 5-gram inputs on the character level. The sequence characters were separated using # symbols as border characters, and the middle word was within parentheses. The system returned the phoneme sequence of the middle word as output (Figure 4). For instance, for the example input below (A), our Transformer-based system would return the output B, which is the phoneme sequence of the word "غذا", meaning food, in *ezafe* form.

A: "گ ف ت # ش ی ر خ ش ک ( غ ذ ا ) ن و ز ا د # ا س ت"

B: q / z a y e

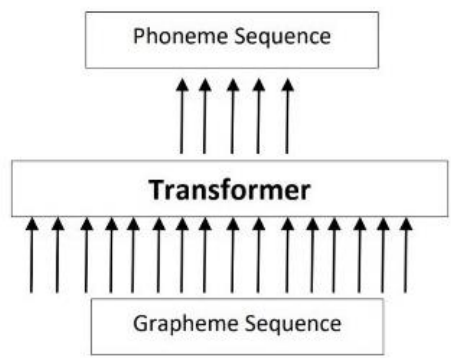

Figure 4. The structure of our end-to-end Transformer-based G2P converter.

## 5. Results

We used accuracy on the word level which calculates the percentage of correctly predicted words by the model that exactly match the reference phoneme sequences over the total number of words. Table 3 shows a comparison of the evaluation results for our experiments. In what follows, we present the performance results of the OOV, *ezafe* and homograph modules in our G2P system.

### 5.1. OOV Model Results

The performance of the OOV model was evaluated on a subset extracted from the original test data with no overlap with the train data. This model achieved a 74.41% accuracy on the word level.

Table 3. Comparison of word-level accuracy between G2P systems.

| System | Word-Level Accuracy |
|---|---|
| Our multi-module system | **94.48%** |
| DNN-based + attention [2], [14] | 87.3% |
| Transformer-based [7] | 91.04% |

### 5.2. *Ezafe* Model Results.

Table 4 indicates a performance comparison between our *ezafe* models. Here, we only report on the word-level accuracy—which is the number of words correctly labeled with *ezafe* over the total number of words—since *ezafe* occurrence was balanced throughout the original test set (unlike the evaluation process carried out in 5.3 for our homograph model). The poor performance of Model II, which is the word-level model, is due to the fact that we are working on low resources; hence the model has few or zero encounters with many words. Here, the character-level approach can help increase the performance through character embeddings.

Model III actually refers to the *ezafe* prediction conducted by our Transformer-based system.

Table 4. Comparison of accuracy between *ezafe* models.

| Model Name | Architecture | Accuracy |
|---|---|---|
| Model I | Character-level GRU-based | **97.14%** |
| Model II | Word-level GRU-based | 82.41% |
| Model III | Component of end-to-end Transformer | 94.32% |

### 5.3. Homograph Model Results

The homograph model was evaluated on a subset extracted from the original test set where the middle gram was a homograph, and achieved a 91.38% word-level accuracy in predicting the phoneme sequence for homographs. However, due to the scatteredness of homographs throughout our data, accuracy is not a good measure to understand the performance of this model. Therefore, using the concept of sensitivity and specificity in statistics, we propose a measure called *homograph score*, which ranges from 0 to 1 with 1 being the best score, and evaluates a model's performance in predicting the correct phoneme sequence associated with homographs.

(1)
$$S_j = \frac{\sum_{i=1}^{n} \frac{\text{number of correct calls for } (P_i) \text{ by the model}}{\text{number of appearances of } (P_i) \text{ among labels throughout test set}}}{n}$$

(2)
$$\text{HomographScore} = \frac{\sum_{i=1}^{C} S_j}{C}$$

(1) calculates $S_j$, which is the sum of all correct calls for pronunciation $P_i$ over the total appearances of $P_i$ in the test dataset, divided by $n$, which is the number of possible pronunciations of homograph $j$. (2) calculates (1) for every unique homograph in the test set, and the summation is divided by the total number of unique homographs in the test set (C), resulting in *homograph score*. Our homograph model attained a *homograph score* of 0.7348.

## 6. Future Work

After the service is released for customer use, we can log the words categorized as OOV and add their phoneme sequences to our look-up table.

Homographs often have less frequent variants. To provide the model with more training examples of the less frequent word(s), we can run our homograph model on raw text and corpora and manually check the predictions assuming the less frequent pronunciations. We can then use the correct predictions to make the data less imbalanced and retrain our homograph model in the hope of attaining a better *homograph score*.

In case a word can be followed by *ezafe*, it is shown in the Ariana Lexicon with the GEN symbol. Currently, every word in the input text is passed to the *ezafe* module. We can use the information from the lexicon to detect words that are never followed by *ezafe* and not pass them to the *ezafe* module, thereby increase system performance.

Since we are working on a low-resource task, an acoustic model using this G2P system would not be able to detect the position of stress properly on its own. To solve this problem, we are currently using the rules explained in [16], but for improvement, we can add a component that uses the output of our G2P system (the phoneme sequence) to determine the position of stress in words; therefore, helping the acoustic model produce more natural and humanlike speech signals. This component can comprise a Transformer encoder block to receive a phoneme sequence and a linear layer to produce 0-1 vectors indicating the presence or absence of stress for the phonemes in the middle word.

Needless to say, using pre-trained models and/or having access to sufficient resources would affect the system's performance. In our case, for instance, a pre-trained BERT would improve our system in terms of *ezafe* recognition.

## 7. Conclusions

This paper presents a sequence-level multi-module framework for G2P conversion of Persian text. The system is comprised of a GRU-based model combined with an attention layer, and a Transformer-based model to tackle OOV, homograph and *ezafe* problems. The models were evaluated using the Bijankhan corpus in terms of word-level accuracy. Moreover, we introduce a new evaluation metric called *homograph score* for disambiguating homograph pronunciation in G2P tools.

## Acknowledgements

This research was supported by Asr Gooyesh Pardaz Co. We thank Khosro Hosseinzadeh and Farokh Kakaei who provided insight and expertise that assisted the research.